\documentclass[runningheads]{llncs}
\usepackage[T1]{fontenc}
\usepackage{graphicx,verbatim}
\usepackage{color}
\usepackage{xcolor}
\definecolor{darkgreen}{RGB}{0,100,0}  
\usepackage[hidelinks,breaklinks, colorlinks, linkcolor=red, citecolor=blue]{hyperref}
\usepackage{amsmath}
\usepackage{amssymb}
\usepackage{dsfont}
\usepackage{pifont}
\usepackage{algorithm}
\usepackage{algorithmic}
\usepackage{booktabs}
\usepackage{tabularx}
\usepackage{multirow}
\usepackage{makecell}
\usepackage{colortbl}
\usepackage[capitalize]{cleveref}
\usepackage{marvosym}

\crefname{section}{Sec.}{Secs.}
\Crefname{section}{Section}{Sections}
\crefname{table}{Tab.}{Tabs.}
\newcommand{\minisection}[1]{\vspace{0.04in} \noindent {\bf #1}\,}

\def\ourmodel{CMSA-Net}
\begin{document}
\title{CMSA-Net: Causal Multi-scale Aggregation with Adaptive Multi-source Reference for Video Polyp Segmentation}
\titlerunning{CMSA-Net}
%
\vspace{-2mm}
\author{
Tong Wang\inst{1,2} \and
Yaolei Qi\inst{1} \and
Siwen Wang\inst{2} \and
Imran Razzak\inst{2} \and
Guanyu Yang\inst{1}\textsuperscript{(\textrm{\Letter})} \and
Yutong Xie\inst{2}\textsuperscript{(\textrm{\Letter})}
}

\authorrunning{T. Wang et al.}

\institute{
Southeast University, China \and
Mohamed bin Zayed University of Artificial Intelligence (MBZUAI), UAE \\
\url{https://github.com/wangtong627/CMSA-Net}
}

  
\maketitle              
\vspace{-2mm}
\begin{abstract}
Video polyp segmentation (VPS) is an important task in computer-aided colonoscopy, as it helps doctors accurately locate and track polyps during examinations. However, VPS remains challenging because polyps often look similar to surrounding mucosa, leading to weak semantic discrimination. In addition, large changes in polyp position and scale across video frames make stable and accurate segmentation difficult.
To address these challenges, we propose a robust VPS framework named \ourmodel{}. The proposed network introduces a Causal Multi-scale Aggregation (CMA) module to effectively gather semantic information from multiple historical frames at different scales. By using causal attention, CMA ensures that temporal feature propagation follows strict time order, which helps reduce noise and improve feature reliability. Furthermore, we design a Dynamic Multi-source Reference (DMR) strategy that adaptively selects informative and reliable reference frames based on semantic separability and prediction confidence. This strategy provides strong multi-frame guidance while keeping the model efficient for real-time inference.
Extensive experiments on the SUN-SEG dataset demonstrate that \ourmodel{} achieves state-of-the-art performance, offering a favorable balance between segmentation accuracy and real-time clinical applicability. 

\end{abstract}

\section{Introduction}
\vspace{-2mm}
Colorectal cancer (CRC) is one of the leading causes of cancer-related death worldwide~\cite{biller2021diagnosis}. Since most CRC cases develop from colorectal polyps, early detection and accurate segmentation during colonoscopy are critically important. However, the miss rate for polyps during colonoscopy remains as high as 25\%, posing a significant clinical challenge~\cite{jiang2023risk}. Therefore, developing effective video polyp segmentation (VPS) methods is of great clinical value for assisting real-time diagnosis and intervention.

\begin{figure}[t]
    \centering
    \includegraphics[width=\linewidth]{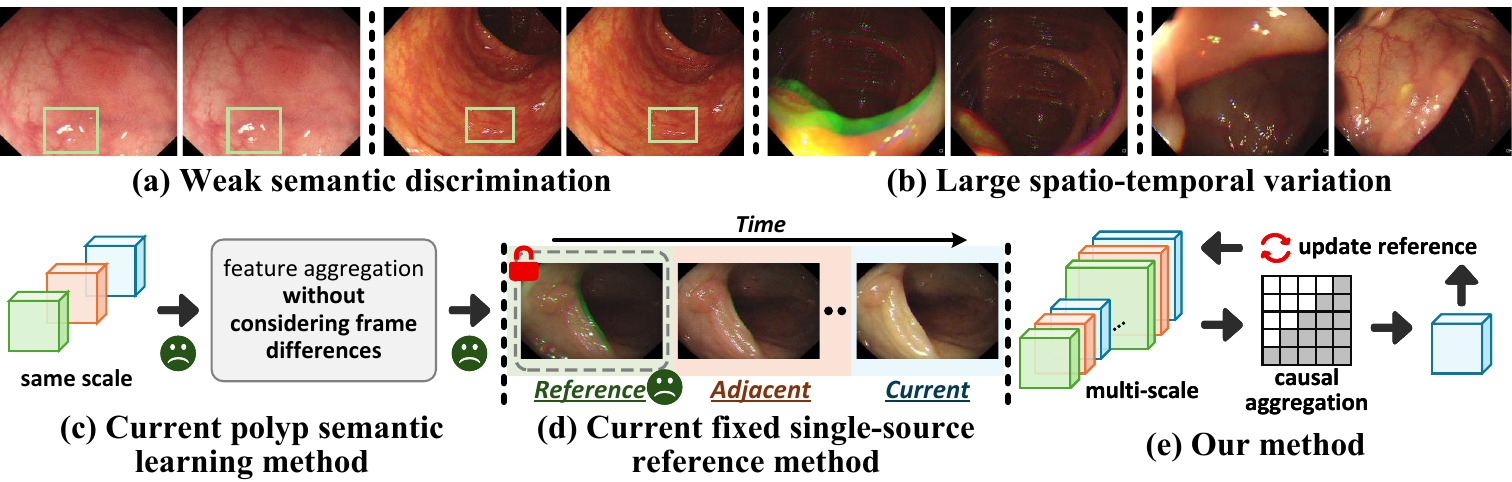}
    \vspace{-4mm}
    \caption{
\textbf{(a) Challenge 1:} Weak semantic discrimination, caused by the low contrast between polyps and background (case14\_6 and case19).
\textbf{(b) Challenge 2:} Large spatio-temporal variations across frames (case51\_1 and case71\_1).
\textbf{(c) Limitation 1:} Existing feature fusion methods for polyp semantic learning , which ignore multi-scale information and intrinsic identity relationships.
\textbf{(d) Limitation 2:} Current approaches adopt a fixed single-source reference, failing to capture dynamic and diverse spatio-temporal cues.
\textbf{(e) Our Method}.
}
    \vspace{-4mm}
    \label{fig:motivation}
\end{figure}

Despite recent progress, accurate VPS still faces challenges (see~\cref{fig:motivation}(a-b)). 
\textbf{(1) Weak semantic discrimination:} polyps often exhibit low semantic contrast with surrounding mucosa, making discriminative semantic learning difficult. 
\textbf{(2) Large spatio-temporal variation:} irregular camera motion causes drastic changes in polyp scale and position across frames, disrupting temporal consistency. 
\textbf{(3) Real-time requirement:} clinical applications demand low-latency inference during procedures.

Early polyp segmentation methods~\cite{fan2020pranet,puyal2020endoscopic,zhang2020adaptive,wei2021shallow,wang2024polyp,wu2022polypseg+,wu2025epsegnet,ye2025fpn,chai2024querynet} followed static image segmentation paradigms and showed limited performance in videos due to the lack of temporal modeling. More recently, VPS methods~\cite{ji2021progressively,ji2022video,chen2024mast,hu2024sali,chen2025stddnet,lu2024diff,zhao2025efficient} introduce additional frames to assist current-frame segmentation. In particular, \textit{reference frames} are used to provide semantic guidance, while \textit{adjacent frames} are exploited to model short-term temporal consistency.
Despite their utilization of temporal information, several limitations still remain:
\textbf{(1) Limited spatio-temporal fusion (see~\cref{fig:motivation}(c)):} existing methods usually aggregate features at a single spatial scale, using Mamba-based modules~\cite{xu2024polyp,fu2025polymamba,chen2025stddnet,zhu2025hybrid} or attention mechanisms~\cite{ji2021progressively,ji2022video,hu2024sali}. This design fails to fully exploit rich semantic cues across different scales in the reference frame and adjacent frames. Moreover, they often overlook the intrinsic identity relationships among the reference, adjacent and current frames, which can lead to feature contamination when large appearance variations occur between frames~\cite{feng2025scoring}.
\textbf{(2) Single-source reference dependency (see~\cref{fig:motivation}(d)):} most approaches rely on a single and fixed reference source, which makes them less robust to large target variations.
While memory-based methods suffer from computational inefficiency due to redundant frames and may fail to retrieve the optimal reference information~\cite{zhou2024rmem}.

To address these issues, we propose the \textbf{C}ausal \textbf{M}ulti-scale \textbf{S}patio-temporal \textbf{A}ggregation \textbf{Net}work (\textbf{\textit{\ourmodel{}}}). \ourmodel{} integrates causal multi-scale modeling with a dynamic multi-source reference strategy to enhance semantic perception of polyps with weak discriminability while maintaining real-time efficiency (see~\cref{fig:motivation}(e)).
(1) We propose the \textbf{Causal Multi-scale Aggregation (CMA)} module for current-frame semantic learning. Unlike single-scale feature interactions, CMA enables the current frame to aggregate multi-scale semantic priors from reference and adjacent frames. By exploiting temporal information across different spatial scales, CMA improves the representation of weakly discriminative polyp features. Moreover, a causal attention mechanism is introduced to model the logical relationships among reference, adjacent, and current frames, encouraging temporally coherent feature propagation and reducing feature contamination under large inter-frame variations.
(2) We propose a \textbf{Dynamic Multi-source Reference (DMR)} strategy to provide multiple meaningful semantic references. DMR adaptively updates the reference set by selecting reliable frames from the video sequence according to the current frame, ensuring semantic consistency while avoiding redundant computation. The resulting multi-source references provide stable and robust semantic cues, which further enhance the effectiveness of CMA and support real-time inference.
We conducted extensive experiments on SUN-SEG dataset. 
Results show that \ourmodel{} achieves state-of-the-art performance while maintaining real-time inference speeds.
Our contributions are summarized as follows:
\begin{itemize}
    \item[$\bullet$] We propose \ourmodel{} for VPS through causal multi-scale modeling and dynamic multi-source references.
    
    \item[$\bullet$] We design a Causal Multi-scale Aggregation (CMA) module for causal multi-scale spatio-temporal aggregation to enhance current-frame semantics.
    
    \item[$\bullet$] We introduce a Dynamic Multi-source Reference (DMR) strategy that selects reliable reference frames for stable and efficient semantic guidance.
    
    \item[$\bullet$] Extensive experiments on the SUN-SEG dataset demonstrate that \ourmodel{} achieves state-of-the-art performance with real-time inference speed.
\end{itemize}

\section{Method}
\begin{figure}[t]
    \centering
    \includegraphics[width=\linewidth]{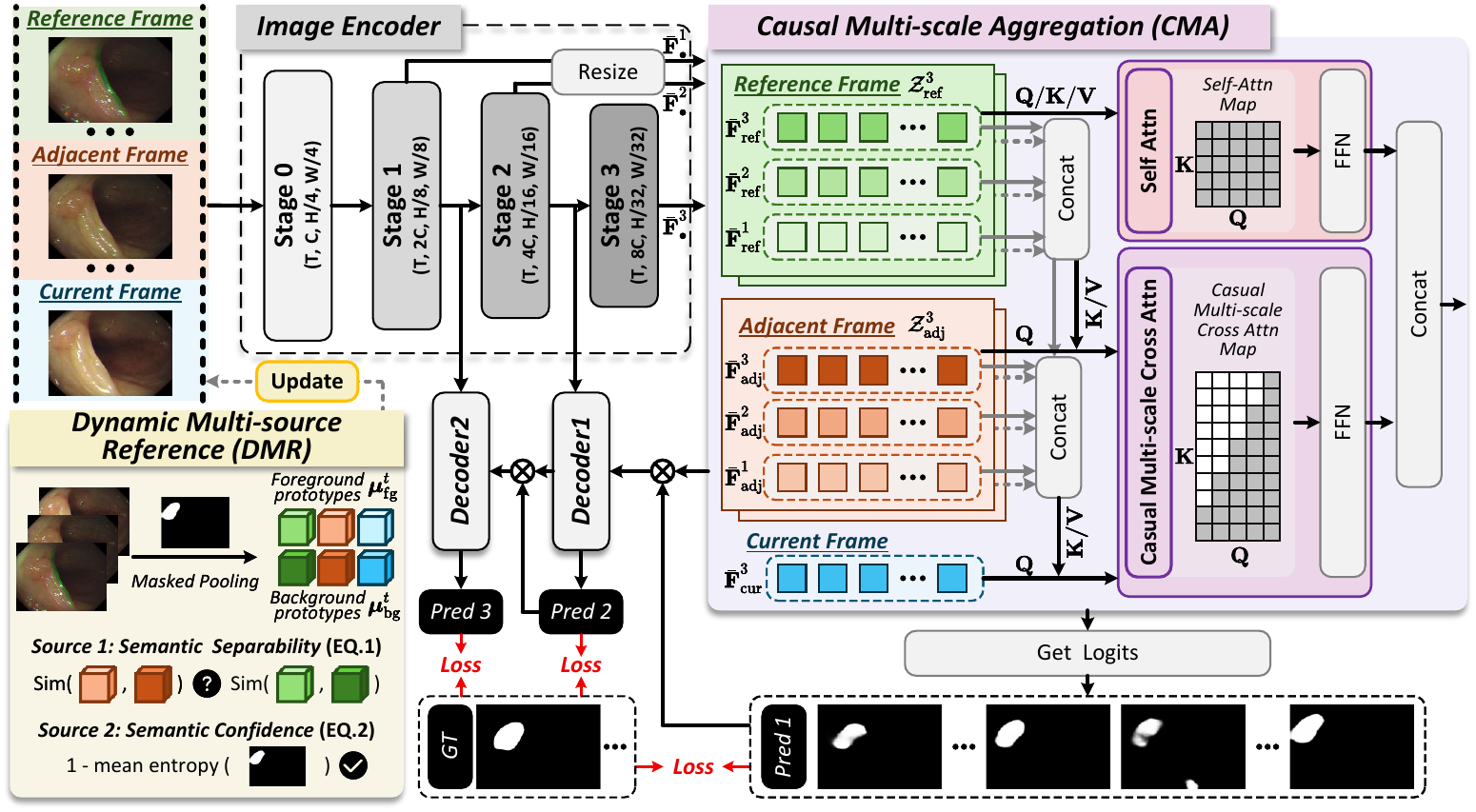}
    \vspace{-6mm}
    \caption{Framework of the proposed \ourmodel{}, including the CMA module
(\cref{subsec:cma}) with DMR strategy (\cref{subsec:dmr}).}
    \vspace{-4mm}
    \label{fig:method}
\end{figure}

\vspace{-2mm}
\cref{fig:method} shows our \ourmodel{}. Given an input video clip with $T$ frames, we denote the clip as $\{I_t\}_{t=1}^{T}$.
The first $R$ frames $\{I_{1}, \dots, I_{R}\}$ are \textit{{reference frames}} $I_\mathrm{ref}$,
where $R$ corresponds to the number of multi-source references;
the last frame $I_T$ is the \textit{{current frame}} $I_\mathrm{cur}$;
and the remaining frames $\{I_{R+1}, \dots, I_{T-1}\}$ are \textit{{adjacent frames}} $I_\mathrm{adj}$.
We extract multi-scale representations for all frames using a shared \textbf{Image Encoder} backbone with $S$ stages.
We denote the feature at the $s$-th stage of the $r$-th reference frame as
$\mathbf{F}^{s}_{\mathrm{ref}_r}$, where $r \in \{1,\dots,R\}$ indexes the reference source and
$s \in \{0,\dots,S\}$ indexes the backbone stage.
Similarly, the $s$-th stage representation of the current frame is denoted as $\mathbf{F}^{s}_{\mathrm{cur}}$,
and that of the $k$-th adjacent frame is denoted as $\mathbf{F}^{s}_{\mathrm{adj}_k}$ for
$k \in \{R+1, \dots, T-1\}$.
For convenience, we concatenate the features of all reference frames, adjacent frames, and the current frame
at the $s$-th stage to form a unified representation:
$
\mathbf{F}^{s} = \big[ \mathbf{F}^{s}_{\mathrm{ref}_1}, \dots, \mathbf{F}^{s}_{\mathrm{ref}_R},
\mathbf{F}^{s}_{\mathrm{adj}_{R+1}}, \dots, \mathbf{F}^{s}_{\mathrm{adj}_{T-1}},
\mathbf{F}^{s}_{\mathrm{cur}} \big],
$
where $[\cdot]$ denotes concatenation.
Following~\cite{ji2022video}, the input clip $\{I_t\}_{t=1}^{T}$ is stacked into $[T, C, H, W]$.
The image encoder produces multi-scale features at 4 stages with decreasing spatial resolutions and increasing channel dimensions, namely,
$\mathbf{F}^{0} \in \mathbb{R}^{T \times 2C \times \tfrac{H}{4} \times \tfrac{W}{4}}$,
$\mathbf{F}^{1} \in \mathbb{R}^{T \times 4C \times \tfrac{H}{8} \times \tfrac{W}{8}}$,
$\mathbf{F}^{2} \in \mathbb{R}^{T \times 8C \times \tfrac{H}{16} \times \tfrac{W}{16}}$,
and
$\mathbf{F}^{3} \in \mathbb{R}^{T \times 16C \times \tfrac{H}{32} \times \tfrac{W}{32}}$.

\subsection{Causal Multi-scale Aggregation (CMA) Module}\label{subsec:cma}
\vspace{-2mm}
CMA enhances the discriminability of current-frame features by aggregating multi-scale spatio-temporal information. For a target backbone stage $s$, the module treats features from all available backbone stages $u \in \{1,\dots,S\}$ as complementary semantic scales to capture rich contextual priors.
Specifically, for any frame $I_\bullet$ in the clip, where the placeholder $\bullet \in \{\text{ref}, \text{adj}, \text{cur}\}$ denotes the frame type, we denote its original feature at the $u$-th stage as $\mathbf{F}^u_\bullet \in \mathbb{R}^{C_u \times H_u \times W_u}$. To enable cross-scale interaction, all source features are aligned to the spatial resolution $(H_s, W_s)$ and unified to the channel dimension $C$ of the target stage $s$ via:
$
    \bar{\mathbf{F}}^{u\rightarrow s}_{\bullet} = \mathrm{Conv}_{1\times1}\!\left(\mathrm{Conv}_{3\times3}\left(\mathrm{Resize}_{H_s\times W_s}(\mathbf{F}^{u}_{\bullet})\right)\right) \in \mathbb{R}^{C \times H_s \times W_s}.
$
These aligned multi-scale features are then concatenated channel-wise to form a per-frame multi-scale token set:
$
    \mathcal{Z}_{\bullet}^s = [\,\bar{\mathbf{F}}^{1\rightarrow s}_{\bullet}, \dots, \bar{\mathbf{F}}^{S\rightarrow s}_{\bullet}\,] \in \mathbb{R}^{(S \cdot C) \times H_s \times W_s}.
$

To leverage temporal priors without future leakage, CMA employs causal attention, restricting features at time $t$ to attend only to reference features and past frames, and accordingly constructs frame-dependent query ($\mathbf{Q}$), key ($\mathbf{K}$), and value ($\mathbf{V}$).
\textbf{(1) For a reference feature} $\bar{\mathbf{F}}_{\mathrm{ref}_r}^{s}$ ($r \in \{1,\dots,R\}$),
self-attention is applied with
$\mathbf{Q}_{\mathrm{ref}_r} = W_Q \cdot \mathcal{Z}_{\mathrm{ref}_r}^s$,
$\mathbf{K}_{\mathrm{ref}_r} = W_K \cdot \mathcal{Z}_{\mathrm{ref}_r}^s$,
and $\mathbf{V}_{\mathrm{ref}_r} = W_V \cdot \mathcal{Z}_{\mathrm{ref}_r}^s$.
\textbf{(2) For an adjacent feature} $\bar{\mathbf{F}}_{\mathrm{adj}_t}^{s}$ ($t \in \{R\!+\!1,\dots,T\!-\!1\}$),
the query is generated from its current-scale feature
$\mathbf{Q}_{t} = W_Q \cdot \bar{\mathbf{F}}_{\mathrm{adj}_t}^{s}$,
while keys and values are constructed by concatenating multi-scale token sets of
all reference frames and adjacent frames up to time $t$:
$\mathbf{K}_{t} = W_K \cdot [\,\mathcal{Z}_{\mathrm{ref}_1}^s,\dots,\mathcal{Z}_{\mathrm{ref}_R}^s,\mathcal{Z}_{\mathrm{adj}_{R+1}}^s,\dots,\mathcal{Z}_{\mathrm{adj}_{t}}^s\,]$
and
$\mathbf{V}_{t} = W_V \cdot [\,\mathcal{Z}_{\mathrm{ref}_1}^s,\dots,\mathcal{Z}_{\mathrm{ref}_R}^s,\mathcal{Z}_{\mathrm{adj}_{R+1}}^s,\dots,\mathcal{Z}_{\mathrm{adj}_{t}}^s\,]$.
\textbf{(3) For the current feature} $\bar{\mathbf{F}}_{\mathrm{cur}}^{s}$ ($t=T$),
the query is generated from its high-level representation
$\mathbf{Q}_{\mathrm{cur}} = W_Q \cdot \bar{\mathbf{F}}_{\mathrm{cur}}^{s}$,
and keys and values are formed by concatenating multi-scale token sets of all available frames:
$\mathbf{K}_{\mathrm{cur}} = W_K \cdot [\,\mathcal{Z}_{\mathrm{ref}_1}^s,\dots,\mathcal{Z}_{\mathrm{ref}_R}^s,\mathcal{Z}_{\mathrm{adj}_{R+1}}^s,\dots,\mathcal{Z}_{\mathrm{adj}_{T-1}}^s,\mathcal{Z}_{\mathrm{cur}}^s\,]$
and
$\mathbf{V}_{\mathrm{cur}} = W_V \cdot [\,\mathcal{Z}_{\mathrm{ref}_1}^s,\dots,\mathcal{Z}_{\mathrm{ref}_R}^s,\mathcal{Z}_{\mathrm{adj}_{R+1}}^s,\dots,\mathcal{Z}_{\mathrm{adj}_{T-1}}^s,\mathcal{Z}_{\mathrm{cur}}^s\,]$.
All attention operations strictly follow the causal constraint, ensuring that no feature at time $t$
attends to features from future frames.
The causal multi-scale aggregation is
$
\mathcal{A}_t^{s}
=
\mathrm{Softmax}\!\left(
\frac{\mathbf{Q}_t \cdot \mathbf{K}_t^{\top}}{\sqrt{d}}
\right)\mathbf{V}_t
$, where $d$ denotes the feature dimension.
The aggregated output is fused with the original feature through residual learning:
$
\hat{\mathbf{F}}^{s}_{t} = \bar{\mathbf{F}}^{s}_{t} + \mathcal{A}_t^{s}, \quad
\tilde{\mathbf{F}}^{s}_{t}
=
\hat{\mathbf{F}}^{s}_{t}
+
\mathtt{FFN}\!\big(\mathtt{LN}(\hat{\mathbf{F}}^{s}_{t})\big)
$. CMA aggregates multi-scale temporal context
to enhance polyp representations.

\subsection{Dynamic Multi-source Reference (DMR) Strategy}\label{subsec:dmr}
\vspace{-2mm}
While CMA aggregates multi-scale spatio-temporal semantics, its effectiveness
relies on high-quality references.
Existing VPS methods depend on fixed references~\cite{chen2025stddnet,ji2021progressively}
or memory-based designs~\cite{hu2024sali,zhou2024rmem},
which are unreliable over time, computationally expensive, and inflexible.
We propose a DMR strategy that adaptively maintains a compact set of references during inference, leveraging two complementary sources: semantic separability and semantic confidence.

\minisection{Semantic Separability–based reference update.}
Given the CMA output feature $\tilde{\mathbf{F}}_t^{s}$ at stage $s$, we first obtain the prediction logits by a $1\times1$ convolution,
and the corresponding probability map $p_t \in [0,1]^{H\times W}$ after sigmoid activation.
The probability map is used to perform masked pooling on $\tilde{\mathbf{F}}_t^{s}$ to extract foreground and background prototypes:
$
\boldsymbol{\mu}_{\mathrm{fg}}^{t}
=
\frac{\sum_{i} p_t(i)\,\mathbf{f}_t(i)}{\sum_{i} p_t(i)}, \quad
\boldsymbol{\mu}_{\mathrm{bg}}^{t}
=
\frac{\sum_{i} (1-p_t(i))\,\mathbf{f}_t(i)}{\sum_{i} (1-p_t(i))}
$, where $\mathbf{f}_t(i)$ denotes the feature vector at spatial location $i$.
For a candidate frame $t$, we evaluate its semantic quality by foreground--background semantic separability and temporal consistency:
\vspace{-2mm}
\begin{equation}
\vspace{-2mm}
s_{\mathrm{sep}}^{t} = 1 - \cos\!\left(\boldsymbol{\mu}_{\mathrm{fg}}^{t}, \boldsymbol{\mu}_{\mathrm{bg}}^{t}\right), \\
s_{\mathrm{cons}}^{t} = \cos\!\left(\boldsymbol{\mu}_{\mathrm{fg}}^{t}, \boldsymbol{\mu}_{\mathrm{fg}}^{\mathrm{cur}}\right), \\
\mathrm{Score}_{\mathrm{sem}}^{t}
=  s_{\mathrm{sep}}^{t} + s_{\mathrm{cons}}^{t}.
\end{equation}
The semantic reference $\tilde{\mathbf{F}}_{\mathrm{ref}}^{s,\mathrm{sem}}$ is updated only
when a higher $\mathrm{Score}_{\mathrm{sem}}$ is achieved, with a cooldown interval to prevent drift.

\minisection{Semantic Confidence–based reference update.}
In parallel, we maintain a confidence reference $\tilde{\mathbf{F}}_{\mathrm{ref}}^{s,\mathrm{conf}}$
to capture frames with reliable predictions.
The semantic confidence of a candidate frame $t$ is quantified by an
entropy-based determinacy measure:
\vspace{-3mm}
\begin{equation}
\vspace{-3mm}
c^{t}
=
1 - \frac{1}{|\Omega|}
\sum_{i\in\Omega}
H\!\left(p_t(i)\right), \quad
\mathrm{Score}_{\mathrm{conf}}^{t}
=
c^{t} + s_{\mathrm{cons}}^{t},
\end{equation}
where $H(p) = -p\log p - (1-p)\log(1-p)$.
The confidence reference $\tilde{\mathbf{F}}_{\mathrm{ref}}^{s,\mathrm{conf}}$ is updated only
when a higher $\mathrm{Score}_{\mathrm{conf}}$ is achieved, with a cooldown interval.

\subsection{Learning Process and Implementation Details}
\vspace{-2mm}
\minisection{Learning Process and Loss Function.}
During training, we supervise predictions at multiple stages using the decoder~\cite{ji2021progressively} to facilitate stable optimization.
The CMA output feature $\tilde{\mathbf{F}}^{s}_{t}$ is first used to generate a coarse prediction $\text{pred}_1$.
The CMA feature is then fused with the stage-2 feature and fed into \textbf{Decoder1}, yielding $\text{pred}_2$.
Subsequently, the output features are combined with the stage-1 feature and passed to
\textbf{Decoder2} to produce the final prediction $\text{pred}_3$.
All predictions are resized to the ground-truth resolution.
We employ Dice loss ($\mathcal{L}_{\text{Dice}}$), weighted IoU loss ($\mathcal{L}_{\text{wIoU}}$),
and weighted BCE loss ($\mathcal{L}_{\text{wBCE}}$) jointly.
The segmentation loss is defined as
$
\mathcal{L}_{\text{seg}}(m) =
\mathcal{L}_{\text{Dice}}(m, g) +
\mathcal{L}_{\text{wIoU}}(m, g) +
\mathcal{L}_{\text{wBCE}}(m, g)
$, and the total training loss is
$
\mathcal{L}_{\text{total}} =
\mathcal{L}_{\text{seg}}(\text{pred}_1) +
\mathcal{L}_{\text{seg}}(\text{pred}_2) +
\mathcal{L}_{\text{seg}}(\text{pred}_3)
$, where $m$ and $g$ denote the predicted masks and the corresponding ground-truth.

\minisection{Implementation Details.}
We implement \ourmodel{} using PyTorch and train it on a single NVIDIA RTX 3090 GPU.
Res2Net-50~\cite{Gao2019Res2NetAN} and PVTv2-B2~\cite{Wang2021PVTVI} are adopted as backbone networks,
both initialized with ImageNet pre-trained weights.
All input frames are resized to $352\times352$, and the batch size is set to 4.
The model is optimized using AdamW with a weight decay of $1\times10^{-4}$ and an initial learning rate of $1\times10^{-4}$,
and trained for 30 epochs.
Each input clip contains 6 frames, including reference, adjacent, and current frames.
In implementation, CMA is applied to the stage-3 feature for aggregating high-level temporal semantics of the current frames, while multi-stage features are still utilized for the reference and adjacent frames.
All intermediate features are projected to a unified channel dimension $C=32$,
and the CMA module is implemented with 4 attention heads.
For the DMR strategy, the cooldown interval is set to 5 for semantic reference updates
and 1 for confidence reference updates.

\section{Experiments}\label{sec:experiments}
\vspace{-2mm}
\subsection{Dataset and Metrics}\label{subsec:datasets}
\vspace{-2mm}
\minisection{Dataset.} 
We evaluate on SUN-SEG~\cite{ji2022video}, the largest VPS dataset. Following~\cite{ji2022video,chen2025stddnet}, the training set contains 112 video clips with 19,544 frames, while the test set is divided into four subsets: SUN-SEG-Easy-Seen (33 clips/4,719 frames), Easy-Unseen (86 clips/12,351 frames), Hard-Seen (17 clips/3,882 frames), and Hard-Unseen (37 clips/8,640 frames). Easy/Hard denote segmentation difficulty, while Seen/Unseen indicate whether test videos overlap with the training videos.
\minisection{Metrics.} 
We employ six metrics: Dice, IoU, MAE, structure-measure ($S_\alpha$)~\cite{fan2017structure}, enhanced-alignment measure ($E_{\phi}^{mn}$)~\cite{fan2021cognitive} and weighted F-measure ($F_\beta^w$)~\cite{margolin2014evaluate}.

\subsection{Comparisons with State-of-the-art Methods}
\vspace{-2mm}
We compare \ourmodel{} with 12 state-of-the-art methods: (1) natural video segmentation (NVS), including COSNet~\cite{lu2019see}, PCSA~\cite{gu2020pyramid}, and 2/3D~\cite{puyal2020endoscopic}; (2) image-based polyp segmentation (IPS), such as PraNet~\cite{fan2020pranet}, ACSNet~\cite{zhang2020adaptive}, SANet~\cite{wei2021shallow}, and SEPNet~\cite{wang2024polyp}; and (3) video-based polyp segmentation (VPS), including PNS~\cite{ji2021progressively}, PNS+~\cite{ji2022video}, MAST~\cite{chen2024mast}, SALI~\cite{hu2024sali}, and STDDNet~\cite{chen2025stddnet}.

\minisection{Quantitative Comparisons.}
\cref{tab:SUN-SEG-Easy} and \cref{tab:SUN-SEG-Hard} present the quantitative comparison results.
\footnote{
For fair comparison, all methods were retrained on the SUN-SEG dataset. As the official SALI only provides a PVTv2-B5 version, we retrained it with the PVTv2-B2. 
}
Overall, \ourmodel{} achieves the best performance across both seen and unseen subsets under different difficulty levels.
On the Easy setting, \ourmodel{} consistently outperforms existing methods on both Seen and Unseen splits, achieving the highest Dice scores and the lowest MAE. On the more challenging Hard setting, our method shows clearer advantages, especially on unseen videos, where it surpasses the strongest baselines by up to \textbf{1.7\%} Dice on Hard-Seen and \textbf{1.1\%} Dice on Hard-Unseen, demonstrating superior robustness and generalization ability. These gains validate the effectiveness of jointly modeling CMA and DMR, where CMA enhances discriminative representations. 
\begin{table}[t]
\centering
\caption{Quantitative comparison on SUN-SEG-Easy dataset. The best is \textbf{bold}.}
\vspace{-2mm}
\renewcommand{\arraystretch}{0.9}
\resizebox{\textwidth}{!}{
\begin{tabular}{l|c|c|cccccc|cccccc}
\toprule
\multirow{2}{*}{Method~} & \multirow{2}{*}{Publication~} & \multirow{2}{*}{Backbone~} & \multicolumn{6}{c|}{SUN-SEG-Easy-Seen (\%)} & \multicolumn{6}{c}{SUN-SEG-Easy-Unseen (\%)} \\
& & & $S_\alpha\uparrow$ & $E_{\phi}^{mn}\uparrow$ & $F_\beta^w\uparrow$ & Dice$\uparrow$ & IoU$\uparrow$ & MAE$\downarrow$ & $S_\alpha\uparrow$ & $E_{\phi}^{mn}\uparrow$ & $F_\beta^w\uparrow$ & Dice$\uparrow$ & IoU$\uparrow$ & MAE$\downarrow$ \\
\midrule
COSNet~\cite{lu2019see} & TPAMI'19 & - & 84.5 & 83.6 & 72.7 & 73.0 & 64.8 & 3.4 & 65.4 & 60.0 & 43.1 & 42.3 & 34.2 & 7.3 \\
PCSA~\cite{gu2020pyramid} & AAAI'20 & - & 85.2 & 83.5 & 68.1 & 70.9 & 60.4 & 3.9 & 68.0 & 66.0 & 45.1 & 45.0 & 35.3 & 7.8 \\
2/3D~\cite{puyal2020endoscopic} & MICCAI'20 & - & 89.5 & 90.9 & 81.9 & 82.9 & 75.6 & 2.1 & 78.6 & 77.7 & 65.2 & 65.6 & 57.0 & 4.4 \\
\midrule
PraNet~\cite{fan2020pranet} & MICCAI'20 & Res2Net-50 & 91.8 & 94.2 & 87.7 & 88.3 & 82.5 & 2.0 & 78.1 & 78.8 & 66.3 & 66.5 & 58.2 & 5.2 \\
ACSNet~\cite{zhang2020adaptive} & MICCAI'20 & ResNet-34 & 92.0 & 94.2 & 87.4 & 88.2 & 82.8 & 1.7 & 77.2 & 76.6 & 63.0 & 63.8 & 56.4 & 4.6 \\
SANet~\cite{wei2021shallow} & MICCAI'21 & Res2Net-50 & 91.6 & 93.3 & 86.6 & 87.2 & 82.0 & 1.8 & 75.0 & 72.8 & 59.0 & 59.3 & 52.4 & 5.2 \\
SEPNet~\cite{wang2024polyp} & TCSVT'24 & PVTv2-B2 & 93.1 & 96.2 & 88.3 & 89.6 & 83.4 & 1.7 & 82.9 & 88.3 & 73.5 & 75.1 & 66.6 & 4.2 \\
\midrule
PNS~\cite{ji2021progressively} & MICCAI'21 & Res2Net-50 & 90.6 & 91.0 & 83.6 & 84.1 & 78.3 & 2.0 & 76.7 & 74.4 & 61.6 & 61.8 & 54.5 & 4.8 \\
PNS+~\cite{ji2022video} & MIR'22 & Res2Net-50 & 91.7 & 92.5 & 84.8 & 85.5 & 78.7 & 2.1 & 80.6 & 79.8 & 67.6 & 67.8 & 59.1 & 4.4 \\
MAST~\cite{chen2024mast} & Arxiv'24 & PVTv2-B2 & 92.5 & 96.2 & 87.8 & 89.3 & 82.7 & 1.6 & 83.2 & 89.4 & 74.9 & 77.0 & 67.4 & 3.7 \\
SALI~\cite{hu2024sali} & MICCAI'24 & PVTv2-B2 & 90.2 & 93.2 & 84.9 & 85.8 & 78.9 & 2.4 & 73.1 & 75.2 & 58.7 & 59.2 & 50.2 & 6.3 \\
SALI~\cite{hu2024sali} & MICCAI'24 & PVTv2-B5 & 90.7 & 93.7 & 85.1 & 86.2 & 79.6 & 2.2 & 77.1 & 82.1 & 64.6 & 65.6 & 56.8 & 5.5 \\
STDDNet~\cite{chen2025stddnet} & ICCV'25 & Res2Net-50 & 93.5 & 96.0 & 89.7 & 90.5 & 85.0 & 1.5 & 81.7 & 83.0 & 72.1 & 72.4 & 64.3 & 3.7 \\
STDDNet~\cite{chen2025stddnet} & ICCV'25 & PVTv2-B2 & 94.1 & 96.9 & 90.5 & 91.5 & 86.1 & 1.4 & 86.0 & \textbf{90.3} & 78.6 & 80.1 & 72.4 & 3.4 \\
\midrule
Ours & - & Res2Net-50 & 94.5 & 97.3 & 90.5 & 91.9 & 86.5 & 1.2 & 84.4 & 90.1 & 75.3 & 77.5 & 69.2 & 3.5 \\
Ours & - & PVTv2-B2 & \textbf{95.1} & \textbf{97.5} & \textbf{91.6} & \textbf{92.6} & \textbf{87.6} & \textbf{1.1} & \textbf{86.7} & \textbf{90.3} & \textbf{79.3} & \textbf{80.3} & \textbf{72.6} & \textbf{2.9} \\
\bottomrule
\end{tabular}
}
\label{tab:SUN-SEG-Easy}
\end{table}

\begin{table}[t]
\centering
\caption{Quantitative comparison on SUN-SEG-Hard dataset. The best is \textbf{bold}.}
\vspace{-2mm}
\renewcommand{\arraystretch}{0.9}
\resizebox{\textwidth}{!}{
\begin{tabular}{l|c|c|cccccc|cccccc}
\toprule
\multirow{2}{*}{Method~} & \multirow{2}{*}{Publication~} & \multirow{2}{*}{Backbone~} & \multicolumn{6}{c|}{SUN-SEG-Hard-Seen (\%)} & \multicolumn{6}{c}{SUN-SEG-Hard-Unseen (\%)} \\
& & & $S_\alpha\uparrow$ & $E_{\phi}^{mn}\uparrow$ & $F_\beta^w\uparrow$ & Dice$\uparrow$ & IoU$\uparrow$ & MAE$\downarrow$ & $S_\alpha\uparrow$ & $E_{\phi}^{mn}\uparrow$ & $F_\beta^w\uparrow$ & Dice$\uparrow$ & IoU$\uparrow$ & MAE$\downarrow$ \\
\midrule
COSNet~\cite{lu2019see} & TPAMI'19 & - & 78.5 & 77.2 & 62.6 & 63.3 & 54.1 & 4.6 & 67.0 & 62.7 & 44.3 & 43.8 & 35.3 & 7.0 \\
PCSA~\cite{gu2020pyramid} & AAAI'20 & - & 77.2 & 75.9 & 56.6 & 58.5 & 47.9 & 5.7 & 68.2 & 66.0 & 44.2 & 45.0 & 35.1 & 8.0 \\
2/3D~\cite{puyal2020endoscopic} & MICCAI'20 & - & 84.9 & 86.9 & 75.3 & 76.4 & 67.1 & 3.5 & 78.6 & 77.5 & 63.4 & 64.4 & 55.8 & 4.4 \\
\midrule
PraNet~\cite{fan2020pranet} & MICCAI'20 & Res2Net-50 & 88.4 & 91.9 & 83.1 & 83.9 & 76.6 & 3.1 & 78.7 & 80.2 & 66.7 & 67.5 & 58.7 & 5.3 \\
ACSNet~\cite{zhang2020adaptive} & MICCAI'20 & ResNet-34 & 87.2 & 91.0 & 80.6 & 82.0 & 74.8 & 3.6 & 76.2 & 77.6 & 61.0 & 62.4 & 54.7 & 5.3 \\
SANet~\cite{wei2021shallow} & MICCAI'21 & Res2Net-50 & 87.4 & 90.5 & 81.0 & 82.0 & 74.8 & 3.3 & 75.3 & 73.6 & 59.0 & 59.5 & 52.7 & 5.5 \\
SEPNet~\cite{wang2024polyp} & TCSVT'24 & PVTv2-B2 & 89.4 & 94.0 & 83.5 & 85.7 & 77.6 & 3.4 & 84.7 & 89.5 & 74.5 & 77.4 & 68.4 & 3.9 \\
\midrule
PNS~\cite{ji2021progressively} & MICCAI'21 & Res2Net-50 & 87.0 & 89.2 & 78.7 & 79.6 & 72.1 & 3.3 & 76.7 & 75.5 & 60.9 & 61.5 & 53.9 & 5.0 \\
PNS+~\cite{ji2022video} & MIR'22 & Res2Net-50 & 88.7 & 90.2 & 80.6 & 81.3 & 72.8 & 3.0 & 79.8 & 79.3 & 65.4 & 66.1 & 57.1 & 5.0 \\
MAST~\cite{chen2024mast} & Arxiv'24 & PVTv2-B2 & 89.2 & 94.2 & 83.2 & 85.3 & 76.7 & 2.6 & 85.6 & 91.3 & 77.2 & 79.9 & 70.8 & 3.1 \\
SALI~\cite{hu2024sali} & MICCAI'24 & PVTv2-B2 & 86.8 & 90.9 & 79.9 & 81.0 & 72.6 & 3.4 & 72.8 & 75.9 & 56.9 & 57.9 & 48.7 & 6.8 \\
SALI~\cite{hu2024sali} & MICCAI'24 & PVTv2-B5 & 86.6 & 91.0 & 79.7 & 81.0 & 72.9 & 3.8 & 76.5 & 81.3 & 62.0 & 63.6 & 54.7 & 5.7 \\
STDDNet~\cite{chen2025stddnet} & ICCV'25 & Res2Net-50 & 91.3 & 95.2 & 86.9 & 88.1 & 81.0 & 2.3 & 83.4 & 85.6 & 74.1 & 75.0 & 67.3 & 3.7 \\
STDDNet~\cite{chen2025stddnet} & ICCV'25 & PVTv2-B2 & 91.1 & 95.0 & 86.0 & 87.8 & 80.6 & 2.8 & 86.3 & 90.2 & 78.1 & 80.2 & 72.2 & 3.5 \\
\midrule
Ours & - & Res2Net-50 & \textbf{92.7} & \textbf{96.1} & \textbf{87.1} & \textbf{89.8} & \textbf{83.0} & \textbf{1.8} & 85.1 & 89.8 & 75.0 & 78.0 & 69.5 & 3.6 \\
Ours & - & PVTv2-B2 & 92.3 & 95.6 & \textbf{87.1} & 88.9 & 82.1 & 1.9 & \textbf{87.3} & \textbf{91.0} & \textbf{79.6} & \textbf{81.3} & \textbf{73.7} & \textbf{2.9} \\
\bottomrule
\end{tabular}
}
\label{tab:SUN-SEG-Hard}
\vspace{-4mm}
\end{table}

\begin{figure}[t]
    \centering
    \includegraphics[width=0.95\linewidth]{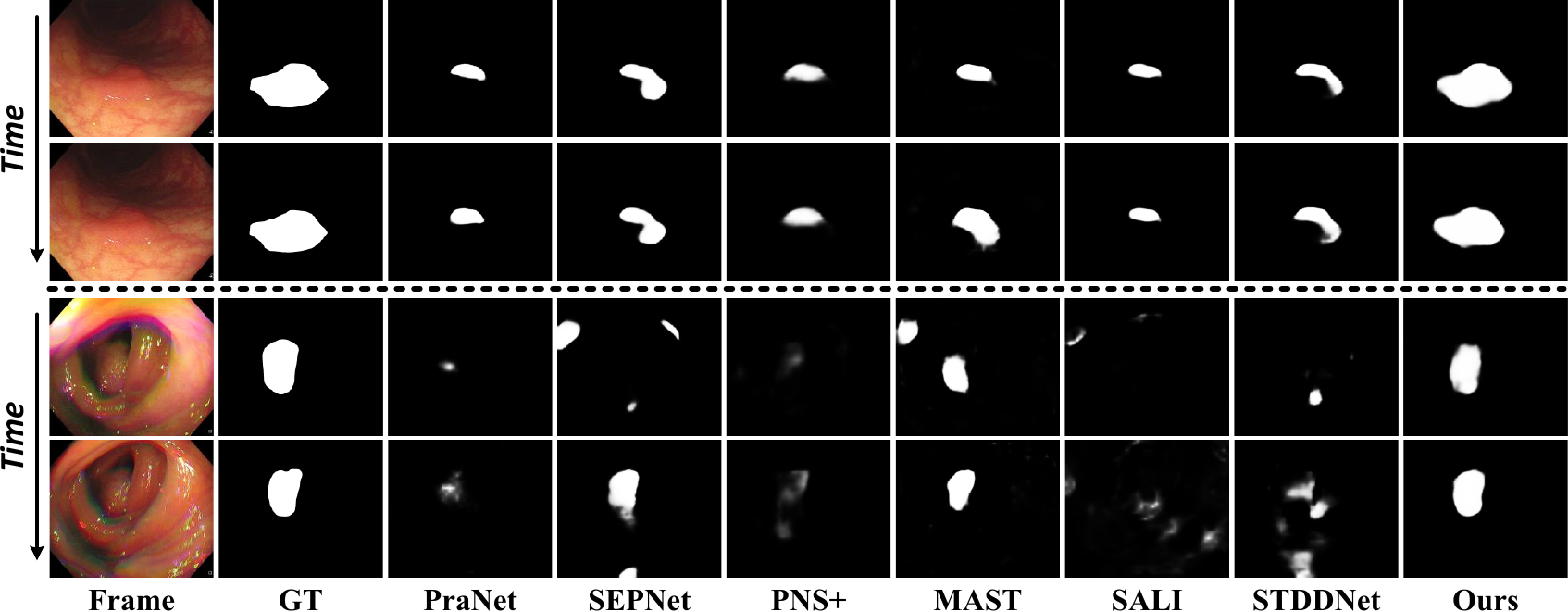}
    \vspace{-3mm}
    \caption{Qualitative Comparisons. Upper (case91\_2): a sequence of
consecutive low-contrast frames; Lower (case32\_4): significant variations between adjacent frames.}
    \label{fig:qualitative_comparisons}
\end{figure}

\begin{table*}[t]
\centering
\caption{Comparison of model parameters and inference time (clip = 6).}
\vspace{-2mm}
\renewcommand{\arraystretch}{0.9}
\resizebox{0.85\textwidth}{!}{
\begin{tabular}{r|c|c|c|c|c|c|c|c|c}
\toprule
Efficiency & PraNet & ACSNet & SEPNet & PNS+ & SALI & STDDNet (R) & STDDNet (P) & Ours (R) & Ours (P)\\
\midrule
GFLOPs $\downarrow$ & 78.89 & 130.52 & 75.12 & 68.58 & 86.58 & 75.17 & 64.18 & 73.36 & 59.42 \\
Param. (M) $\downarrow$ & 30.50 & 29.45 & 25.96 & 9.8 & 82.73 & 38.16 & 29.67 & 29.61 & 25.79 \\
FPS $\uparrow$ & 42 & 21 & 33 & 64 & 10 & 45 & 36 & 47 & 38 \\
\bottomrule
\end{tabular}
\label{tab:model_complexity}
}
\end{table*}
\begin{table}[!ht]
    \centering
    \caption{Ablation study of key components. The best is \textbf{bold}.}
    \vspace{-2mm}
    \renewcommand{\arraystretch}{1}
    \resizebox{\textwidth}{!}{
    \begin{tabular}{l|ccccc|ccccc|ccccc|ccccc}
        \toprule
        \multirow{2}{*}{Methods} 
        & \multicolumn{5}{c|}{SUN-SEG-Easy-Seen} 
        & \multicolumn{5}{c|}{SUN-SEG-Easy-Unseen} 
        & \multicolumn{5}{c|}{SUN-SEG-Hard-Seen} 
        & \multicolumn{5}{c}{SUN-SEG-Hard-Unseen} \\
        & $S_\alpha$ & $E_{\phi}^{mn}$ & $F_\beta^w$ & Dice & IoU
        & $S_\alpha$ & $E_{\phi}^{mn}$ & $F_\beta^w$ & Dice & IoU
        & $S_\alpha$ & $E_{\phi}^{mn}$ & $F_\beta^w$ & Dice & IoU
        & $S_\alpha$ & $E_{\phi}^{mn}$ & $F_\beta^w$ & Dice & IoU \\
        \midrule
        \ourmodel{}
        & \textbf{95.1} & \textbf{97.5} & \textbf{91.6} & \textbf{92.6} & \textbf{87.6}
        & \textbf{86.7} & \textbf{90.3} & \textbf{79.3} & \textbf{80.3} & \textbf{72.6}
        & \textbf{92.3} & \textbf{95.6} & \textbf{87.1} & \textbf{88.9} & \textbf{82.1}
        & \textbf{87.3} & \textbf{91.0} & \textbf{79.6} & \textbf{81.3} & \textbf{73.7} \\
        
        \midrule
        w/o CMA
        & 92.4 & 94.8 & 87.2 & 88.5 & 82.1
        & 78.1 & 79.5 & 64.4 & 65.2 & 56.5
        & 88.5 & 92.1 & 81.3 & 83.1 & 74.9
        & 76.4 & 78.2 & 61.8 & 62.9 & 54.4 \\
        w/o DMR
        & 93.8 & 96.2 & 89.1 & 90.3 & 84.7
        & 80.2 & 81.6 & 67.5 & 68.8 & 60.3
        & 90.4 & 94.2 & 84.8 & 86.6 & 79.1
        & 79.1 & 80.2 & 65.3 & 67.0 & 58.4 \\
        w/o CMA+DMR
        & 90.6 & 92.9 & 84.0 & 85.3 & 79.0
        & 72.5 & 72.1 & 55.1 & 55.8 & 47.6
        & 85.8 & 89.5 & 77.7 & 79.2 & 71.0
        & 71.6 & 71.5 & 53.4 & 54.8 & 46.3 \\
        \midrule
        w/o Multi-scale
        & 93.2 & 95.8 & 88.6 & 89.7 & 83.9
        & 83.5 & 86.9 & 75.1 & 76.2 & 67.4
        & 89.8 & 93.5 & 84.2 & 86.1 & 78.4
        & 82.9 & 87.5 & 74.1 & 75.8 & 67.1 \\
        w/o Casual Attn
        & 94.2 & 96.9 & 90.1 & 91.3 & 85.6
        & 84.5 & 88.2 & 76.8 & 78.1 & 69.8
        & 91.2 & 95.2 & 85.9 & 87.7 & 80.5
        & 85.3 & 89.4 & 76.7 & 78.5 & 70.2 \\
        \midrule
        w/o Multi-source
        & 94.0 & 96.7 & 89.9 & 91.1 & 85.5
        & 86.7 & 90.7 & 79.7 & 80.9 & 73.0
        & 91.0 & 95.0 & 85.6 & 87.5 & 80.3
        & 86.9 & 89.9 & 78.4 & 80.4 & 72.6 \\
        \bottomrule
    \end{tabular}
    \label{tab:ablation_study}
    }
\end{table}

\minisection{Qualitative Comparisons.}
In \cref{fig:qualitative_comparisons}, we visualize the segmentation results of different methods on challenging cases. Our method achieved the best results.

\minisection{Model Complexity.}
In \cref{tab:model_complexity}, while improving segmentation performance, our method also meets the requirements for real-time inference.

\subsection{Ablation Study}
\vspace{-1mm}
We conduct ablation studies on SUN-SEG to evaluate the effectiveness of each component, with results summarized in \cref{tab:ablation_study}.
\textbf{(1) CMA and DMR.}
Removing CMA or DMR consistently degrades performance, with more severe drops on Hard and Unseen subsets.
On SUN-SEG-Hard-Unseen, the Dice score decreases from 81.3\% to 62.9\% without CMA and to 67.0\% without DMR.
Removing both modules causes a drastic collapse to 54.8\% Dice, indicating that causal temporal aggregation
and dynamic reference modeling are essential and complementary.
\textbf{(2) Multi-scale and Causal Design in CMA.}
Using a single-scale interaction or replacing causal attention with standard cross-attention leads to clear
performance degradation, demonstrating the importance of multi-scale aggregation and causal modeling
for robust temporal representation.
\textbf{(3) Multi-source Reference.}
Using only a single reference also results in inferior performance, validating the benefit of maintaining multiple complementary reference sources.

\section{Conclusion}
\vspace{-2mm}
We presented \ourmodel{}, a causal multi-scale spatio-temporal framework for VPS. By integrating CMA and DMR, \ourmodel{} enhances semantic discrimination under challenging conditions. Extensive experiments on SUN-SEG demonstrate consistent superiority over state-of-the-art methods on both seen and unseen subsets, particularly in hard scenarios, while maintaining real-time efficiency, highlighting its practical value for clinical applications.

\bibliographystyle{splncs04}
\bibliography{mybibliography}

\newpage
\appendix


\section{More Qualitative Results}
\vspace{-4mm}
\begin{figure}[!ht]
    \centering
    \includegraphics[width=0.95\linewidth]{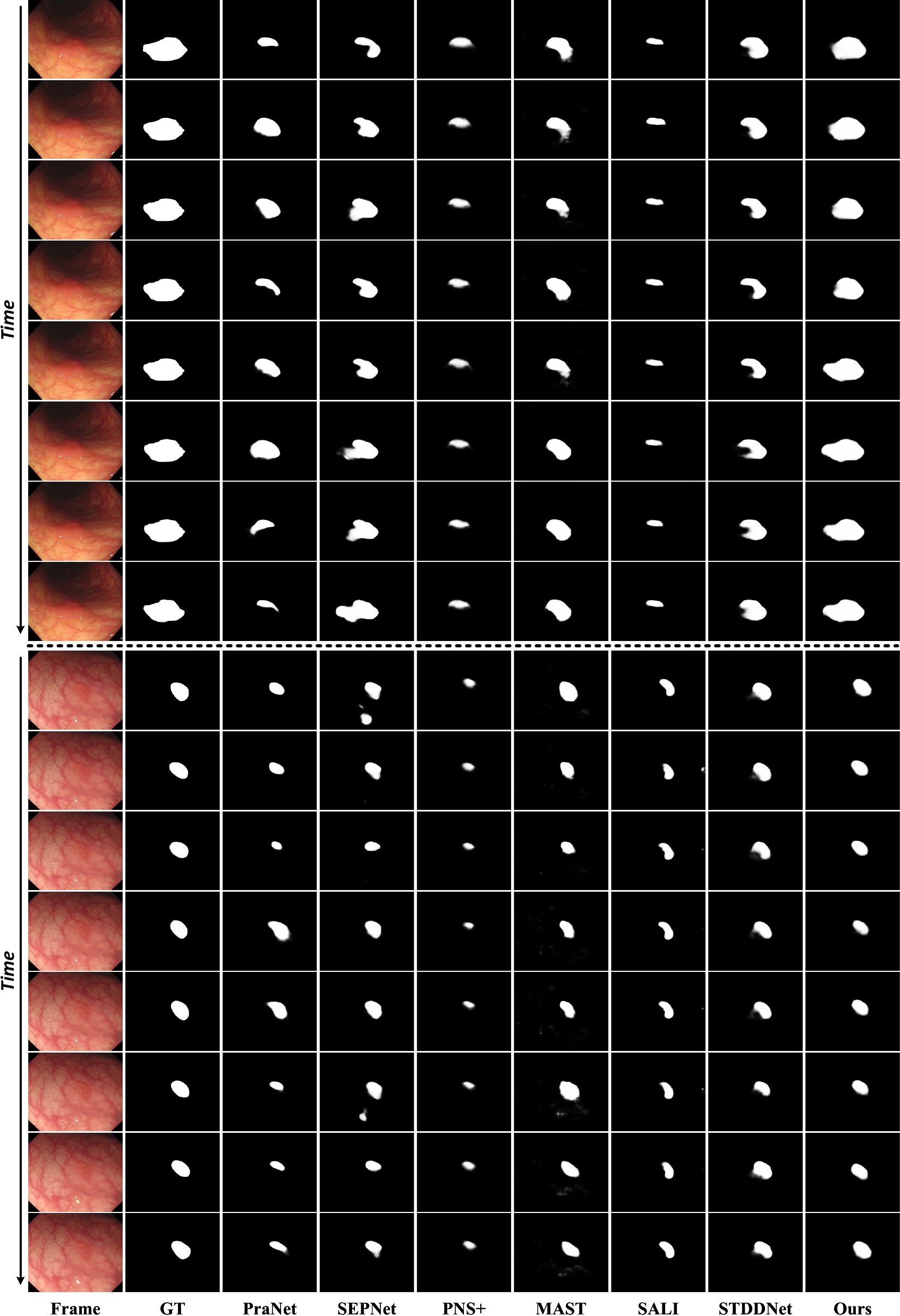}
    \vspace{-3mm}
    \caption{Qualitative Comparisons on consecutive low-contrast frames (case91\_2 and case14\_6).}
    \vspace{-1mm}
    \label{fig:sup_qualitative_comparisons_1}
\end{figure}

\begin{figure}[!ht]
    \centering
    \includegraphics[width=0.95\linewidth]{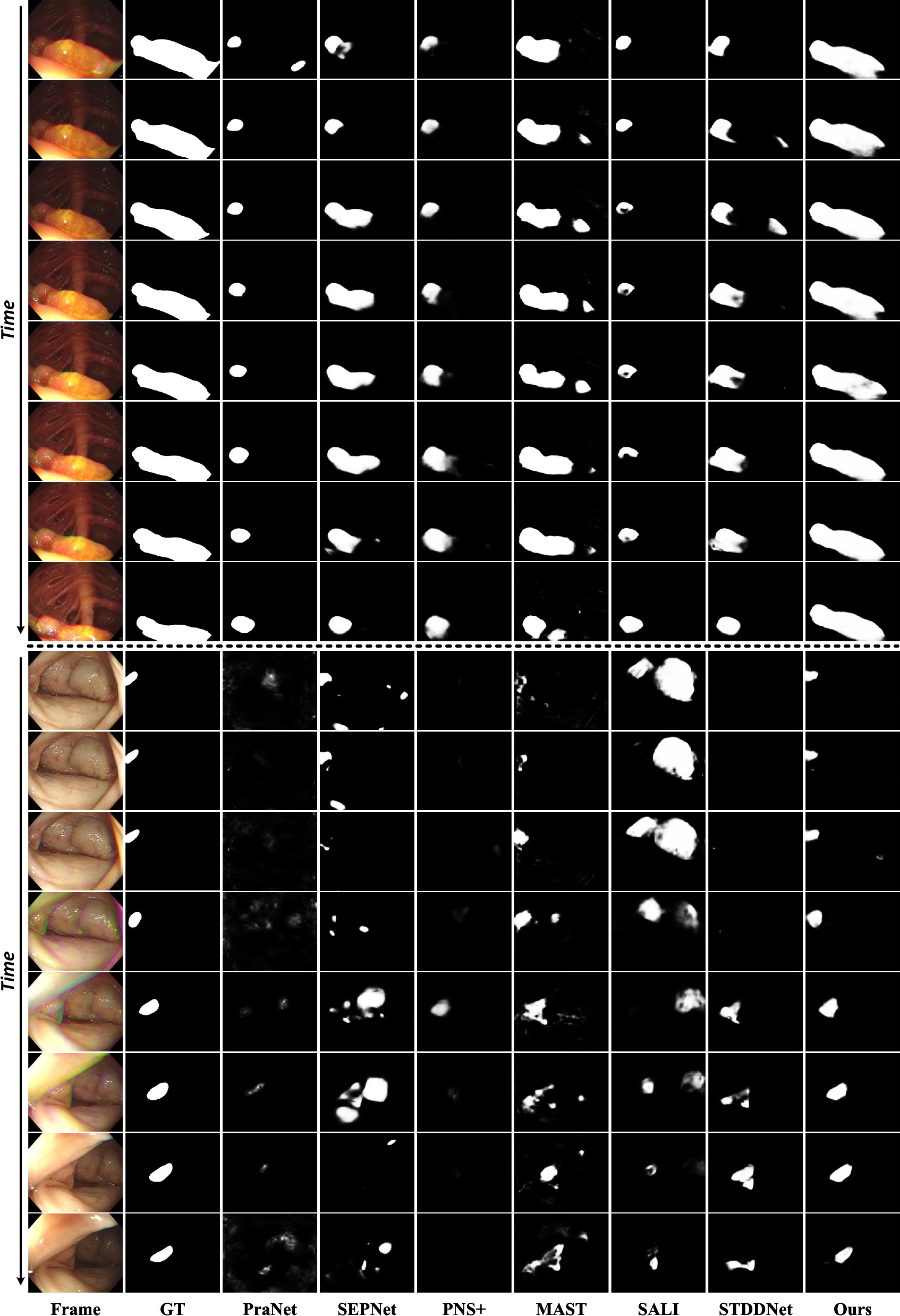}
    \vspace{-3mm}
    \caption{Qualitative Comparisons on consecutive low-contrast and significant-variation frames (case51\_1 and case74\_1).}
    \vspace{-1mm}
    \label{fig:sup_qualitative_comparisons_2}
\end{figure}
\end{document}